\documentclass[preprint,12pt,authoryear]{elsarticle}

\usepackage{amssymb}
\usepackage[authoryear]{natbib}
\usepackage{subcaption}

\usepackage{tikz}
\usepackage{float}
\usepackage[edges]{forest}
\usepackage[T1]{fontenc}
\usepackage{lmodern}  
\usepackage[ruled,vlined]{algorithm2e}
\usepackage[table]{xcolor}
\usepackage{makecell}
\usepackage{multirow}
\usepackage{booktabs}
\usepackage{amsmath}
\usepackage{url}
\usepackage{gensymb}
\usepackage{graphicx}
\usepackage{verbatim}
\usepackage{tabularx} 
\usepackage{algorithmicx}
\usepackage{algpseudocode}
\usepackage{minted}
\usepackage{listings}
\usepackage[utf8]{inputenc}
\RequirePackage{stfloats}
\usepackage{xurl}
\usepackage[colorlinks=true, linkcolor=blue, citecolor=blue, urlcolor=blue, hidelinks]{hyperref} 
\usepackage{adjustbox}
\usepackage{siunitx}

\def\tsc#1{\csdef{#1}{\textsc{\lowercase{#1}}\xspace}}
\tsc{WGM}
\tsc{QE}

\begin{document}

\begin{frontmatter}


\title{Active perception for robotic harvesting: 3D reconstruction and localisation of tomatoes hidden within clusters in a Mediterranean greenhouse}


\author[UAL-Inf]{Fernando Cañadas-Aránega}\corref{corr}\ead{fernando.ca@ual.es}
\cortext[corr]{Corresponding author}

\author[Cy]{R. Border}\ead{border.rowan@ucy.ac.cy}

\author[UAL-Inf]{José C. Moreno}\ead{jcmoreno@ual.es}

\author[UAL-Eng]{José L. Blanco-Claraco}\ead{jlblanco@ual.es}


\affiliation[UAL-Inf]{
    organization={Department of Informatics, CIESOL, ceiA3, Universidad de Almería},
    addressline={Ctra. Sacramento s/n},
    city={Almería},
    postcode={04120},
    country={Spain}
}

\affiliation[UAL-Eng]{
    organization={Department of Engineering, CIESOL, ceiA3, Universidad de Almería},
    addressline={Ctra. Sacramento s/n},
    city={Almería},
    postcode={04120},
    country={Spain}
}

\affiliation[Cy]{
    organization={Vision for Robotics Lab (V4RL), University of Cyprus},
    addressline={allipoleos 75},
    city={Nicosia},
    postcode={1678},
    country={Cyprus}
}


\begin{abstract}
Automating robotic harvesting in intensive agriculture within Mediterranean greenhouses requires overcoming significant challenges related to the geometric complexity of plants and occluded fruits. Although existing literature offers solutions targeting crops that grow in isolation (e.g., apples, sweet peppers, or peaches), the fundamental challenge lies in cluster-growing vegetables, where fixed sensors mounted on robotic systems fail to detect fruits hidden behind the visible surface. To address this limitation, this study presents a comprehensive pipeline for the 3D reconstruction and precise localization of each fruit within a cluster, including heavily occluded instances. The proposed methodology is structured into five sequential stages: i) point cloud acquisition using the AgriSEE Next Best View (NBV) active planner; ii) stochastic noise filtering via Statistical Outlier Removal (SOR); iii) surface classification and segmentation using Region Growing (RG); iv) isolation and recovery of occluded fruits through Density-Based Spatial Clustering of Applications with Noise (DBSCAN); and v) 3D pose estimation (position and orientation). This approach extracts the complete cluster geometry, ensuring the reliable identification of partially hidden tomatoes. Evaluated across multiple scenarios with varying occlusion levels within a simulation framework rigorously validated against real-world conditions, the system achieves a precision exceeding 90\%, an average recall of 82.8\%, and a mean Intersection over Union (mIoU) of 80.7\%. Furthermore, it demonstrates high repeatability in centroid estimation with a Root Mean Square Error (RMSE) of merely 4.2~mm, verifying its technical feasibility and high accuracy for autonomous harvesting operations.
\end{abstract}



\end{frontmatter}

\section{Introduction} 
Currently, the global area dedicated to greenhouse cultivation exceeds 490,000 hectares, with an estimated annual growth rate of 19\% since 1980. Of this area, approximately 20\% is located in the southeastern Iberian Peninsula, where the area allocated to greenhouse vegetable farming exceeds 77,000 hectares \citep{parvej2025review}. In Spain, the predominant model corresponds to the style greenhouse, which accounts for 92\% of the global surface area of this type of structure and is characterized by a low-to-medium technological level \citep{sanchez2024robotics}. However, this production model faces growing competition both from highly technified systems, such as those developed in the Netherlands, and from low-cost alternatives implemented in countries like Morocco or Turkey. In this context, enhancing crop productivity and quality is essential to maintaining sector competitiveness \citep{udekwe2025virtual}. Compounding this challenge are the rising global demand for food, the progressive scarcity of rural labor, and the growing interest in autonomous systems—accelerated in part by the COVID-19 pandemic—all of which underscore the need to explore advanced robotic solutions for protected agriculture \citep{ko}.

Although greenhouses exhibit a certain degree of structural organization, their nature is far from that of highly controlled industrial environments \citep{amin2026design,canadas2024greenbot}. This distinction introduces specific challenges that demand the incorporation of machinery with a high level of automation to ensure the efficient operation of greenhouse agriculture. A key aspect in integrating robotic systems into these environments is the selection of suitable sensors capable of detecting obstacles and localizing the robot in dynamic, occluded scenarios with communication constraints \citep{canadas2024autonomous,cao2026global}. These systems must not only navigate autonomously but also execute complex tasks such as automated fruit harvesting, where spatial precision and three-Dimensional (3D) perception play a decisive role \citep{kang2025predicting,ushiroji2026automatic}.

This drive is motivated by the pursuit of improved greenhouse efficiency, ranging from maximizing available internal cultivation space to minimizing costs associated with operational tasks. During the growing season, an estimated 80\% of operational time is devoted to crop monitoring and fruit harvesting, the latter being one of the most widely researched tasks \citep{moreno2024feasibility}. Consequently, the most successful developments in this domain have incorporated robotic arms to automate harvesting, typically equipped with RGB-D cameras for shape and depth detection, and/or LiDAR sensors for 3D environment scanning and mapping \citep{ambrus2024field}, although these sensors are also used for navigation between aisles in greenhouses to make use of their depth-sensing technology and carry out ground segmentation \citep{canadas2026greenseg}. However, although camera-based or LiDAR sensors have mainly been used for navigation, person identification or multi-agent target tracking \citep{canadas2026integrated}, automated fruit harvesting using traditional sensors presents significant hurdles that have limited its large-scale adoption. 2D sensors, such as conventional RGB cameras, are highly sensitive to illumination variations, shadows cast by foliage, and chromatic shifts caused by fruit ripening stages, which degrades the robustness of detection algorithms \citep{chen2026review}.

On the one hand, if we focus our study on the types of sensors used, numerous studies have addressed fruit localization in greenhouse environments using fixed RGB or RGB-D cameras. \citep{yang2024development} employed a stereo camera mounted on a robotic arm to estimate the 3D position of tomato fruits from a fixed viewpoint, combining a 2D detection network with depth triangulation. Similarly, \citep{zhen2026real} used an RGB-D sensor fixed to the chassis of a mobile platform to detect and localize strawberry, relying on a single depth frame per plant. \citep{zhang2026tomato} proposed a multi-camera fixed array to increase angular coverage around the tomato plant, reducing—but not eliminating—the impact of occlusions on detection rate. Other works have explored LiDAR-based sensing for fruit and plant structure localization. \citep{usenko2025using} used a fixed 3D LiDAR to scan tomato leafs rows and segment individual trusses based on point density and geometric features, while \citep{ci20243d} integrated a solid-state LiDAR mounted at a static height to estimate fruit position through clustering of the raw point cloud. \citep{kang2022accurate} combined a fixed RGB camera with a LiDAR sensor to fuse color and depth information for fruit detection in apple orchards, reporting improved localization accuracy over single-sensor configurations. Across these approaches, the sensor—whether camera, LiDAR, or a fixed combination of both—remains statically positioned relative to the plant during acquisition, capturing a single viewpoint or a fixed set of viewpoints per fruit or cluster. This acquisition strategy has proven effective in crops with relatively open canopy structures, but its capacity to resolve fruits located behind other plant elements is inherently limited by the number and placement of the fixed sensors, rather than by the sensing modality itself. This makes the automated harvesting of fruit grown in clusters, such as tomatoes, extremely difficult; this is currently an unresolved problem that requires a solution to ensure production in greenhouses.

On the other hand, if we focus on developing algorithms for fruit detection, precise fruit segmentation and localization using such sensors remain an open challenge due to the geometric variability of plants and fruits. Methods such as $K$-means require prior knowledge of the cluster count and assume simple geometries, whereas approaches based on Random Sample Consensus (RANSAC) rely on parametric geometric models ill-suited for complex organic objects \citep{rajesh2011application, zhou2021design}. Density-based techniques, such as Density-Based Spatial Clustering of Applications with Noise (DBSCAN), are appealing in unstructured environments, but their performance is sensitive to the point cloud density variations inherent to greenhouse acquisitions \citep{ambrus2024field}. Complementarily, region-growing methods exploit local surface continuity, yet they remain vulnerable to noise and vegetation-induced discontinuities \citep{li2025process}. To address these issues, combining scanning techniques with robotic arms, it is possible to go beyond what fixed sensors can observe within their field of view.  That is why the combination of robotic arms and point-cloud scanning techniques offers a potential solution for those fruits that are hidden. Next Best View (NBV) reconstruction with hybrid segmentation algorithms emerges as a robust solution for identifying truss-bound fruits. This combination yields reliable spatial data, essential for enabling automated harvesting tasks and adaptive crop yield estimation \citep{border2018surface}. To address these challenges, Next Best View (NBV) planning algorithms offer a promising solution by incrementally acquiring multiple observations to individually characterize each fruit. These approaches enable the isolated segmentation and reconstruction of each fruit, even under severe occlusions, and accurately determine its 3D spatial coordinates. This information is critical not only for enabling the robotic arm to plan efficient manipulation trajectories based on reliable geometric relations, but also for conducting precise fruit counts to estimate crop yield across a harvesting season \citep{aranega2025nbv}.

This work presents a significant advancement of the Agriculture Surface Edge Explorer (AgriSEE) system introduced in \citep{aranega2025nbv}, expanding it with a comprehensive framework for 3D reconstruction, filtering, segmentation, and pose localization of fruit clusters tailored for automated greenhouse harvesting. The primary objective is to analyze the 3D reconstruction of a tomato plant with fruit clusters using SEE++, the NBV planning algorithm described in \citep{border2018surface} and \citep{border2024surface}, based on point clouds acquired via localized multi-view scans. On the reconstructed model, a noise and outlier removal process is applied using the Statistical Outlier Removal (SOR) algorithm. Following filtering, a hybrid segmentation approach combining region growing and density-based clustering is proposed to isolate individual fruits and estimate their centroids and spatial orientation, thereby yielding their estimated 3D poses. The experimental results demonstrate that the AgriSEE algorithm, integrated with the proposed segmentation pipeline, generates high-quality 3D observations even under severe occlusions, enabling accurate identification and pose estimation of individual fruits. To conduct the experiments, 11 tomato cluster models validated in \citep{greensys2025} were employed. This validation was carried out under real field conditions by comparing the structure, size, and composition against a dataset of 3000 images \citep{canadas2024greenbot}. Although this work relies on simulation, this validation confirms its direct transferability to field applications, as the harvesting algorithm and robotic arm kinematics have already been real-world tested in \citep{border2024surface}. For each of these simulations, a spatial modeling framework is presented to characterize each detected tomato and estimate its pose, identifying its growth orientation to facilitate subsequent robotic harvesting. This approach establishes a solid foundation for automated harvesting systems and yield estimation in cluster-growing crops, proposing a viable solution for real greenhouse environments.

The remainder of this paper is organized as follows: Section~\ref{sec 2: Mat y met} describes the experimental infrastructure, the proposed methodology, and the simulation setup. Section~\ref{Sec 3: res} presents the experimental results obtained from the simulated scenarios. Finally, Section~\ref{Sec 4: conc} summarizes the main conclusions of this research.

\section{Materials and Methods} \label{sec 2: Mat y met}

This section describes the materials and methods employed in the simulation experiments presented in this work.

\subsection{Materials}

This subsection details the primary components utilized during the simulation execution.

\subsubsection{Tomato Cluster Model}

\begin{figure}[H]
  \centering
  \begin{subfigure}{0.9\linewidth} \centering
    \includegraphics[width=8cm]{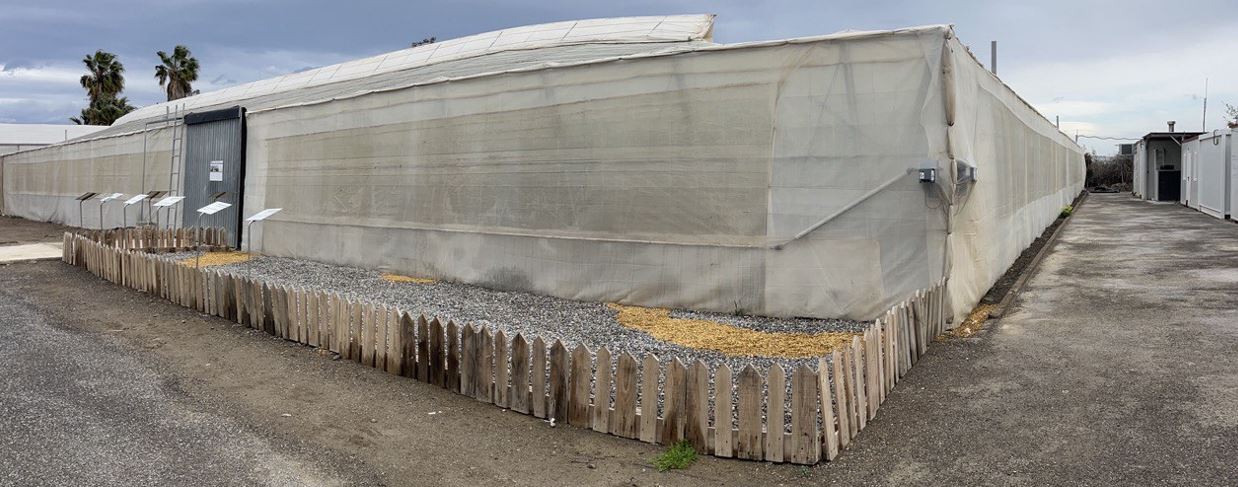} \centering
    \caption{Exterior}
    \label{fig:sub1}
  \end{subfigure}
  \begin{subfigure}{0.9\linewidth} \centering
    \includegraphics[width=8cm]{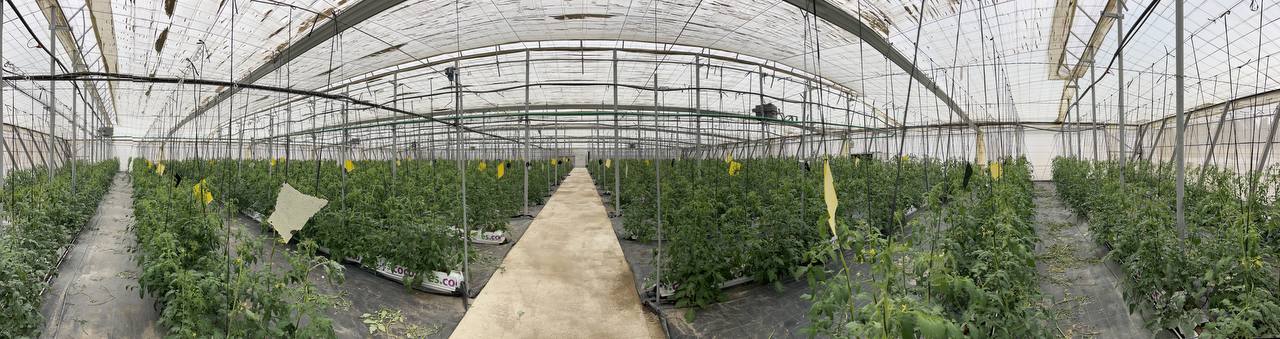} \centering
    \caption{Interior}
    \label{fig:sub2}
  \end{subfigure}
  \caption{AgroConnect experimental greenhouse} \label{fig:Fig1}
\end{figure}

For the simulation, a model of a tomato plant cultivated in the experimental greenhouse belonging to the AgroConnect project (Fig.~\ref{fig:Fig1}), located in La Cañada de San Urbano (Almería, Spain), was employed. The facility corresponds to a Mediterranean-style greenhouse, widely deployed across the southeastern Iberian Peninsula and ideal for crops such as tomatoes and peppers. The interior features a 2-m-wide central aisle acting as the primary access axis, from which eleven side aisles branch out on each side where the target plants are located. The aisles situated in the northern zone have a width of 2~m and a length of 12.5~m, whereas those in the southern zone maintain the same width but reach a length of 22.5~m.

\begin{figure}[H]
  \centering
  \begin{subfigure}{0.45\linewidth}\centering
    \includegraphics[width=4cm]{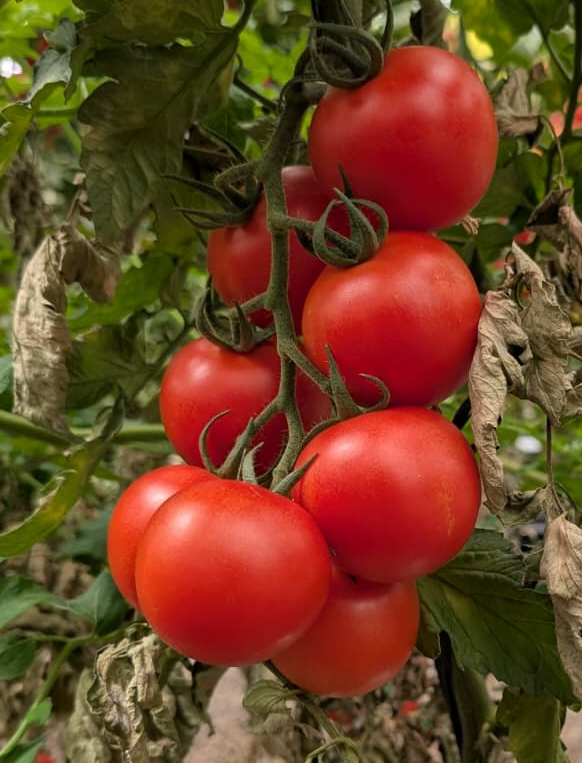}
    \caption{Real tomato cluster}\centering
    \label{fig:real_toma}
  \end{subfigure}
  \begin{subfigure}{0.45\linewidth}\centering
    \includegraphics[width=4cm]{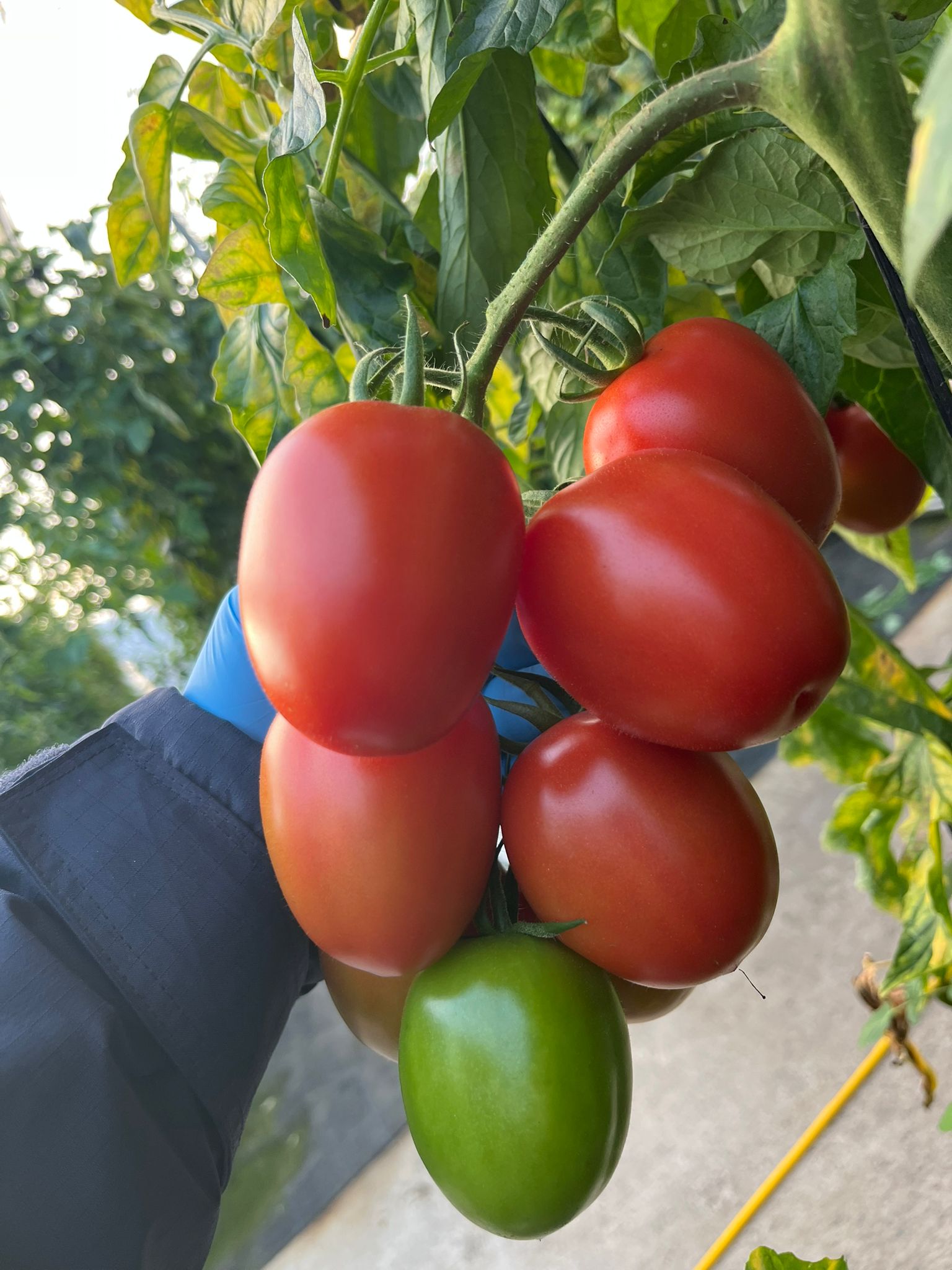}
    \caption{3D cluster model}
    \label{fig:caballera222}
  \end{subfigure}
  \caption{Real tomatoes cluster}
  \label{fig:Figt}
\end{figure}

Inside the greenhouse, the current crop grows organized in trusses comprising multiple tomatoes, ranging from 6 to 9 units per cluster (Fig.~\ref{fig:Figt}). 

\begin{figure}[H]
\centering
  \includegraphics[width=12cm]{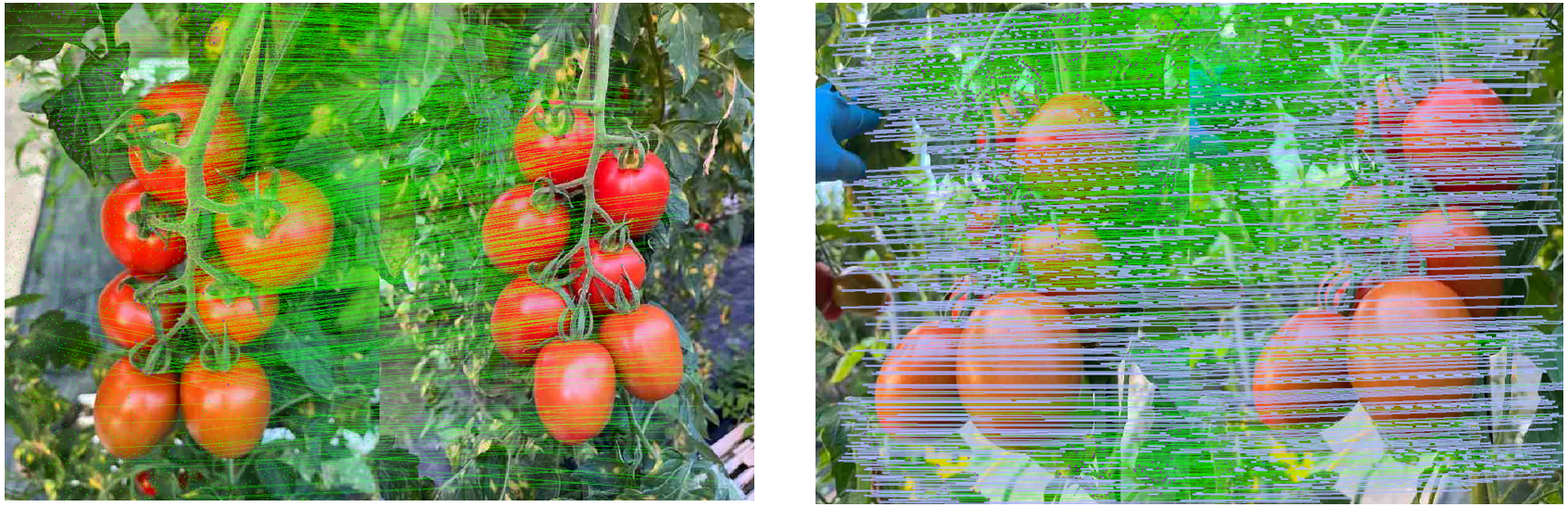}\centering
  \caption{Matching found that coincide between different frames}\label{fig222}
\end{figure}

To evaluate the proposed process, in \citep{greensys2025} we proposed obtaining models of the tomatoes under real-world conditions using the GLOMAP mapper, based on the "Structure-from-Motion`` technique, in conjunction with the Hierarchical Localisation (HLoc) algorithm (Fig. \ref{fig222}). 

Using this technique, real tomato clusters with varying fruit counts were scanned and reconstructed in 3D, refined via SOR filtering and RANSAC-based shape fitting. The resulting models were validated against ground-truth measurements (size, relative position, occlusion, and orientation) using real scanned tomatoes showed in Fig.~\ref{fig:Figt}. To reduce computational overhead, simulations considered a single plant with 11 distinct cluster topologies. This topic is discussed in greater detail in section \ref{Sec 3: res}.

\begin{figure}[H]
  \centering
  \begin{subfigure}{0.45\linewidth}\centering
    \includegraphics[width=4cm]{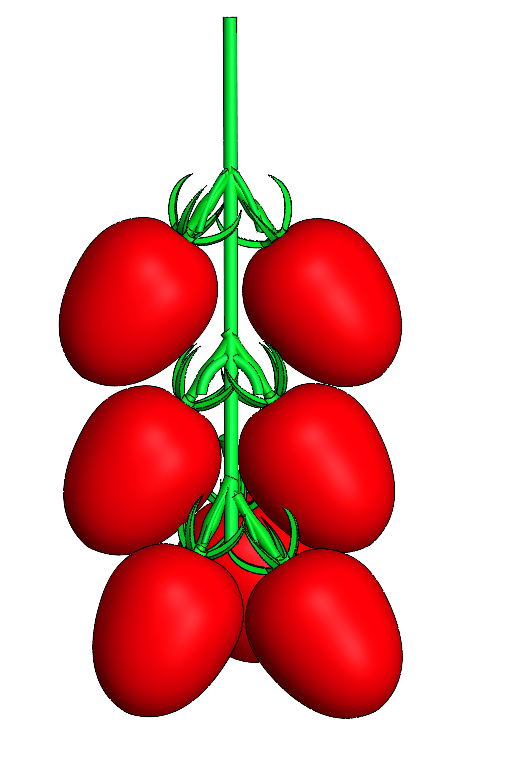}
    \caption{Front view simulated tomato}
    \label{fig:real_toma2}
  \end{subfigure}
  \begin{subfigure}{0.45\linewidth}\centering
    \includegraphics[width=4cm]{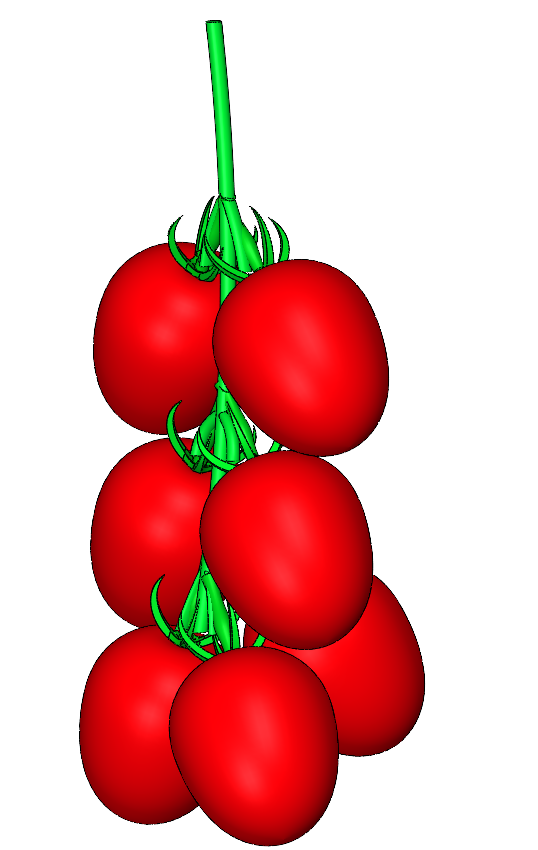}
    \caption{Side view simulated tomato}
    \label{fig:caballera22}
  \end{subfigure}
  \caption{Simulated tomatoes cluster}
  \label{fig:Figt2}
\end{figure}

\subsubsection{AGI Mobile Robot and UR3 Robotic Arm Model}

\begin{figure}[H]
\centering
  \includegraphics[width=5cm]{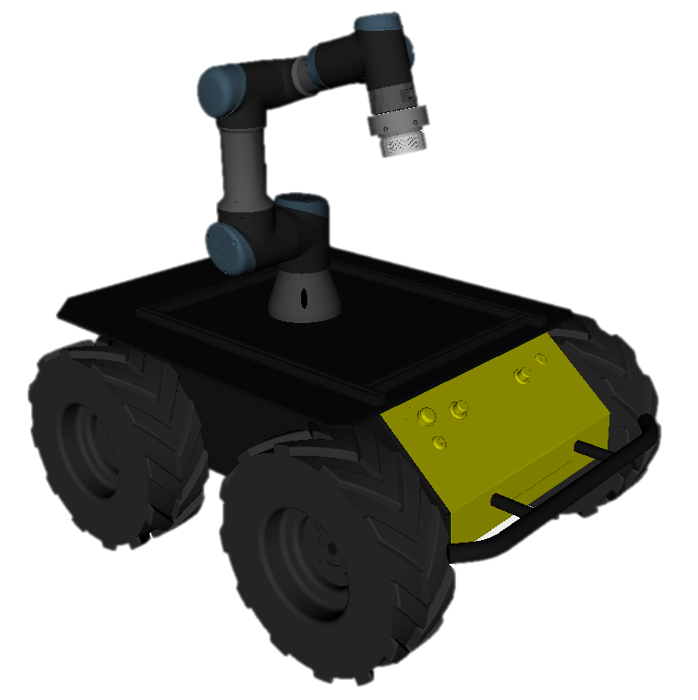}\centering
  \caption{3D model of the AGI mobile robot equipped with the UR3 robotic arm}\label{fig1}
\end{figure}

For the simulations, the AGI mobile robot \cite{Canadasifac2026} (Fig.~\ref{fig1}) was utilized, based on the commercial Husky platform, widely adopted in research due to its robustness and reliability in unstructured environments. This platform has been specifically customized for greenhouse operations by incorporating a custom frame that facilitates autonomous crate transport \cite{moreno2022modelado}. In the simulated configuration, the mobile robot is equipped with a Universal Robots UR3 robotic manipulator, enabling active interaction with the environment. This combination offers high versatility for executing inspection tasks, 3D scanning, and precise manipulation in confined spaces, such as greenhouse aisles.

\subsubsection{Intel RealSense L515 Camera Model}

\begin{figure}[H]
\centering
  \includegraphics[width=6.5cm]{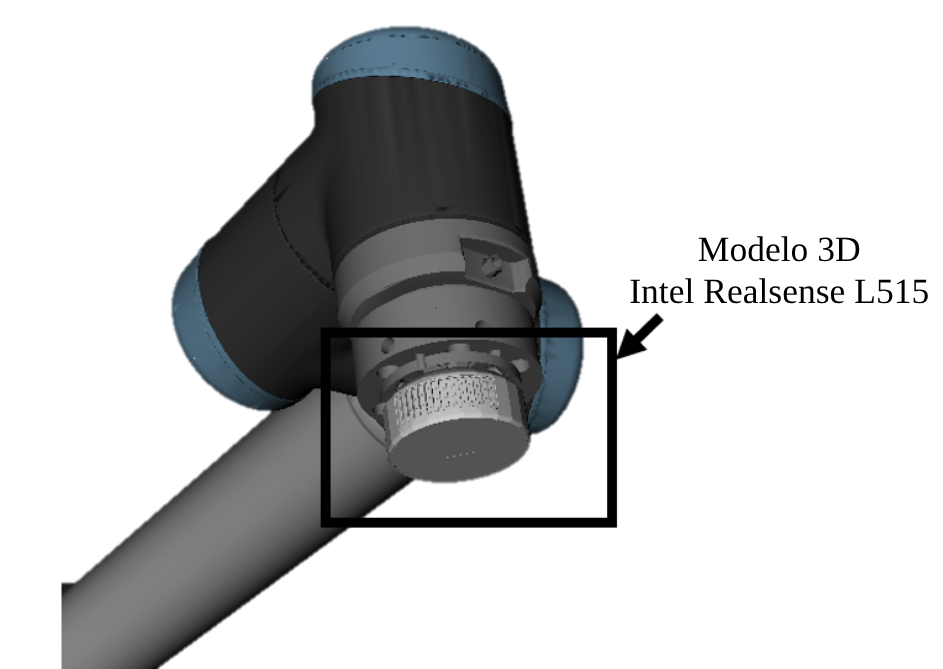}\centering
  \caption{3D model of the Intel RealSense L515 camera}\label{fig2}
\end{figure}

An Intel RealSense L515 camera (Fig.~\ref{fig2}) was mounted on the end-effector of the robotic arm, utilizing solid-state LiDAR technology to acquire depth information during manipulator motion. This sensor provides high resolution in both depth and RGB channels, making it particularly suitable for indoor 3D scanning applications. Data acquisition and synchronization were managed via the Intel RealSense SDK 2.0 and integrated with the Robot Operating System (ROS) Noetic framework, enabling the capture of localized point clouds from multiple viewpoints.

\subsection{Methods}

This subsection describes the methodologies employed during the execution of the simulation experiments.

\subsubsection{Agricultural Surface Edge Explorer (AgriSEE)}

\begin{figure}[H]
    \centering
    \includegraphics[width=7cm]{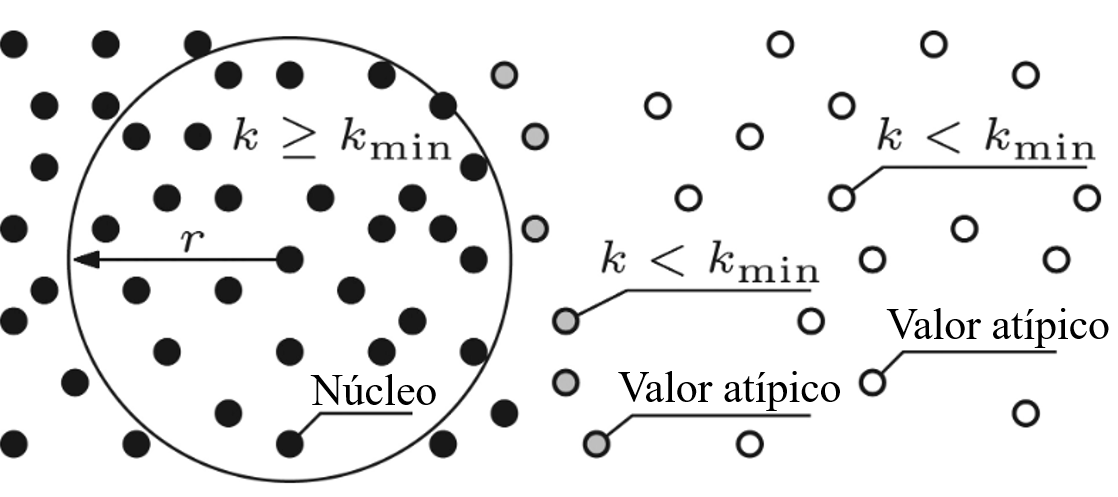}
    \caption{Density-based point classification utilized by AgriSEE \citep{border2018surface}}
    \label{fig:placeholder2}
\end{figure}

AgriSEE is a measurement-guided Next Best View (NBV) planning framework based on the Surface Edge Explorer (SEE++) algorithm \citep{border2024surface}, designed to ensure sufficient coverage of the target surface through iterative selection of informative viewpoints. The algorithm operates directly on sensor-acquired data and relies on local point cloud density analysis, requiring no prior geometric model of the object. Let $\mathcal{P}^k = \{\mathbf{p}_i \in \mathbb{R}^3\}$ be the point cloud acquired at iteration $k$, as illustrated in Fig.~\ref{fig:placeholder2}. For each point $\mathbf{p}_i$, a local density metric $\rho_i^k \in \mathbb{R}^+$ is defined as the number of neighboring points contained within a spherical neighborhood of radius $r \in \mathbb{R}^+$, as expressed in (\ref{eq:density}):
\begin{equation}
\rho_i^k =
\left|
\left\{
\mathbf{p}_j \in \mathcal{P}^k \; \big| \;
\lVert \mathbf{p}_j - \mathbf{p}_i \rVert \leq r
\right\}
\right|.
\label{eq:density}
\end{equation}

Based on this density metric, points are categorized into core points ($\mathbf{p}_c \in \mathbb{R}^3$), frontier points ($\mathbf{p}_f \in \mathbb{R}^3$), or outlier points ($\mathbf{p}_o \in \mathbb{R}^3$), following the criteria proposed in \citep{border2024surface}. Frontier points, characterized by intermediate density, mark transition regions between adequately observed areas and unexplored regions, forming the primary targets for planning subsequent views. For each frontier point $\mathbf{p}_f$, the local surface patch with minimal point density is identified, and the surface normal vector $\mathbf{e}_{\mathrm{n}} \in \mathbb{R}^3$ at $\mathbf{p}_f$ is estimated. A frontier point is considered valid only if its normal points outward from the reconstructed object, which is verified by the condition:
\begin{equation}
\mathbf{e}_{\mathrm{n}} \cdot (\mathbf{p}_f - \mathbf{c}_o) > 0,
\label{eq:normal_condition}
\end{equation}

where $\mathbf{c}_o \in \mathbb{R}^3$ represents the estimated geometric center of the reconstructed object (computed as the arithmetic mean of all accumulated points), and $\cdot$ denotes the dot product. From these valid frontier points, AgriSEE generates candidate viewpoints $\mathbf{v}_k \in \mathbb{R}^3$ by displacing the sensor by a fixed distance $d \in \mathbb{R}^+$ along the normal direction:
\begin{equation}
\mathbf{v}_k = \mathbf{p}_f + d \, \mathbf{e}_{\mathrm{n}}.
\end{equation}

\begin{figure}[H]
    \centering
    \includegraphics[width=6cm]{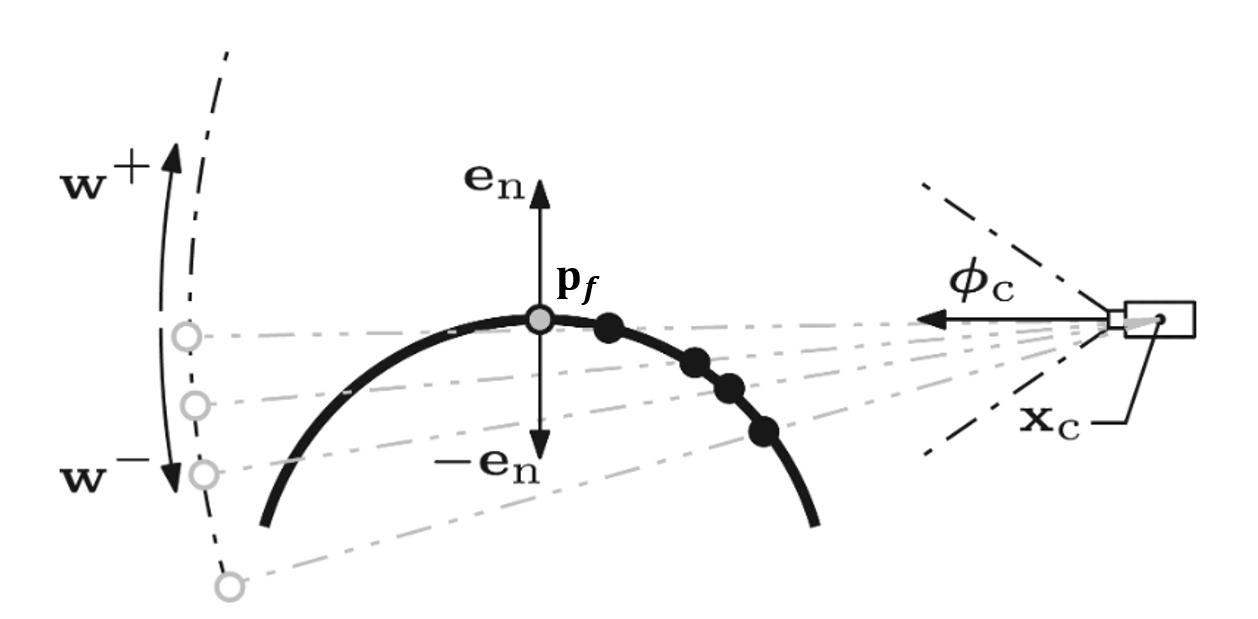}
    \caption{Determination of the correct surface normal orientation \citep{border2018surface}}
    \label{fig:placeholder5}
\end{figure}

This geometric configuration is illustrated in Fig.~\ref{fig:placeholder5}. Here, $\mathbf{w}^+$ and $\mathbf{w}^-$ represent directions derived from the sequential projection of local surface points, used to resolve the correct orientation of the normal vector $\mathbf{e}_{\mathrm{n}}$. Furthermore, the parameters $(\phi_c, x_c)$ describe the observation angle and sensor position relative to the frontier point $\mathbf{p}_f$, respectively, defining the pose of the candidate viewpoint.

The scanning process terminates when a stopping criterion based on a global minimum density threshold is met. Specifically, the reconstruction is deemed sufficiently complete when:
\begin{equation}
\min_{\mathbf{p}_i \in \mathcal{P}^k} \rho_i^k \geq \rho_{\min},
\label{eq:3}
\end{equation}

where $\rho_{\min} \in \mathbb{R}^+$ is a predefined threshold. Once this condition is satisfied, the point clouds acquired across different viewpoints are integrated and refined using an Iterative Closest Point (ICP) registration pipeline, yielding a consolidated 3D model of the cluster.

\subsubsection{Outlier and Noise Removal via Statistical Outlier Removal (SOR)}

To mitigate noise and spurious artifacts, the SOR algorithm was implemented. This method analyzes the spatial distance distribution of each point to its local neighborhood to discriminate noisy elements that do not conform to the underlying surface topology. First, for each point $\mathbf{p}_i$, the mean distance $d_i$ to its $K$-nearest neighbors is computed according to:
\begin{equation}
    d_i = \frac{1}{K} \sum_{j=1}^{K} \text{dist}(\mathbf{p}_i, v_j).
\end{equation}

Next, assuming a Gaussian distribution of spatial distances across the point cloud ($N$), the global mean ($\mu$) and standard deviation ($\sigma$) are determined:
\begin{equation}
    \mu = \frac{1}{N} \sum_{i=1}^{N} d_i,
\end{equation}
\begin{equation}
    \sigma = \sqrt{\frac{1}{N} \sum_{i=1}^{N} (d_i - \mu)^2}.
\end{equation}

An adaptive threshold ($\text{T}_h$) is established based on a standard deviation multiplier $\alpha$, set to $\pm 2\sigma$ (such that approximately 95.4\% of expected inliers lie within the defined boundary):
\begin{equation}
    \text{T}_h = \mu + \alpha \cdot \sigma.
\end{equation}

Consequently, each point in the point cloud is evaluated and classified according to the following decision rule:
\begin{equation}
    P_{\text{filtered}}(\mathbf{p}_i) = 
    \begin{cases} 
        \text{Inlier (Retain)} & \text{if } d_i \le \text{T}_h, \\
        \text{Outlier (Discard)} & \text{if } d_i > \text{T}_h.
    \end{cases}
\end{equation}

By parameterizing the algorithm under a normal distribution, the retention of 95\% of valid geometric data is probabilistically guaranteed, striking an optimal balance between peripheral noise removal and structural preservation. This process effectively removes distant spurious points and noise artifacts arising from stems, leaves, and surrounding foliage.

\subsubsection{Segmentation via Region Growing (RG)}

The subsequent segmentation phase employs a surface-oriented region-growing algorithm applied to the reconstructed and SOR-filtered point cloud. Given the reconstructed point cloud $\mathcal{P} = \{\mathbf{p}_i\}$, where each point $\mathbf{p}_i$ is associated with a unit normal $\mathbf{n}_i \in \mathbb{R}^3$, the angle $\alpha_{ij} \in \mathbb{R}^+$ between the normals of two neighboring points $\mathbf{p}_i$ and $\mathbf{p}_j$ is defined as:
\begin{equation}
\alpha_{ij} = \arccos \left( \mathbf{n}_i \cdot \mathbf{n}_j \right).
\end{equation}

Each unvisited point acts as a seed for a candidate region $R_k \subset \mathcal{P}$, which expands iteratively by incorporating neighboring points $\mathbf{p}_j$ that simultaneously satisfy the following spatial and angular constraints:
\begin{equation}
\lVert \mathbf{p}_j - \mathbf{p}_i \rVert < r_v,
\end{equation}
\begin{equation}
\alpha_{ij} < \theta_{\text{th}},
\end{equation}

where $r_v \in \mathbb{R}^+$ denotes the spatial neighborhood radius and $\theta_{\mathrm{th}} \in \mathbb{R}^+$ represents the maximum allowable angular threshold between normals. This dual criterion ensures that only spatially connected points belonging to a locally smooth surface are grouped, preventing the merging of adjacent instances across curvature discontinuities. A candidate region is accepted as a valid instance only if its cardinality exceeds a minimum threshold:
\begin{equation}
|R_k| \geq N_{\text{min}},
\end{equation}

where $N_{\text{min}} \in \mathbb{N}^+$ serves to discard small fragments and residual noise resulting from partial reconstructions.

\begin{figure}[H]
    \centering
    \includegraphics[width=\linewidth]{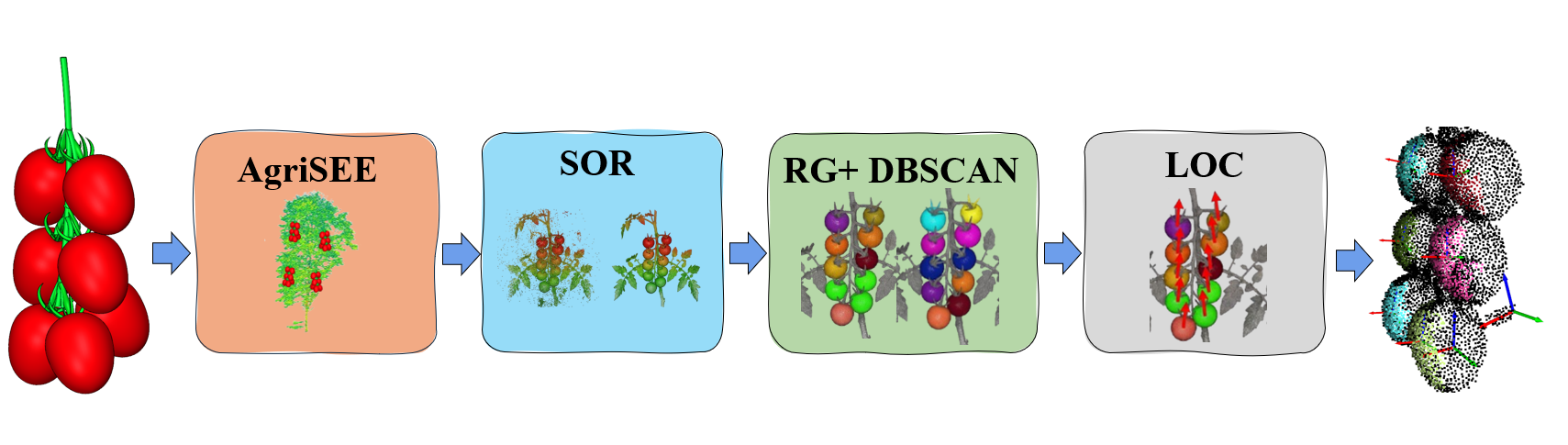}
    \caption{Sequential workflow of the proposed pipeline: from AgriSEE active data acquisition, filtering of outliners points (SOR), segmentation (RG+DBSCAN) adn 3D localization (LOC).}
    \label{fig:sec}
\end{figure}

\subsubsection{Density-Based Spatial Clustering of Applications with Noise (DBSCAN)}

Due to the partial nature of NBV reconstructions, certain fruit instances may not be fully segmented during the Region Growing stage, particularly in heavily occluded cluster regions. To address this limitation, a post-processing stage based on density-based clustering via DBSCAN is incorporated, applied exclusively to the set of unassigned points remaining after primary segmentation \citep{khan2014dbscan}:
\begin{equation}
\mathcal{U} = \left\{ \mathbf{p}_i \in \mathcal{P} \mid \mathbf{p}_i \notin \bigcup_k R_k \right\}.
\end{equation}

DBSCAN is applied to this unassigned subset $\mathcal{U}$ to discover clusters without requiring prior knowledge of the instance count. A point $\mathbf{p}_i \in \mathcal{U}$ is identified as a core point if:
\begin{equation}
\left| \mathcal{N}_{\varepsilon}(\mathbf{p}_i) \right| \geq N_{\min}^{\mathrm{DB}},
\end{equation}

where $\mathcal{N}_{\varepsilon}(\mathbf{p}_i)$ represents the neighborhood of points contained within a sphere of radius $\varepsilon \in \mathbb{R}^+$ centered at $\mathbf{p}_i$. The resulting clusters $n_c \in \mathbb{N}^+$ that meet this criterion are integrated as additional fruit instances, successfully recovering partially observed or hidden tomatoes. Applying DBSCAN strictly to unassigned points prevents interference with previously validated instances and avoids unwanted object merging, delivering a robust final segmentation of the cluster. Fig.~\ref{fig:sec} illustrates the complete sequence executed by the pipeline from active data capture to 3D localization.

\subsubsection{ROS and RViz Integration}

The robotic system framework was developed using ROS Noetic on Ubuntu 20.04 LTS ($x86\_64$ architecture). ROS provided the communication infrastructure required between nodes, enabling seamless integration of the UR3 robotic arm, the Intel RealSense L515 depth sensor, and the point cloud processing modules within the simulation environment. The architecture utilizes custom topics, services, and message types to ensure real-time synchronization between robot kinematics and visual/depth data streams \citep{mishenin2024implementation}.

System visualization and experimental monitoring were conducted using RViz. The Unified Robot Description Format (URDF) model of the UR3 manipulator and real-time point cloud streams were loaded into RViz, enabling visual tracking of arm trajectories and scanning progress. Furthermore, visual markers were utilized to represent planned paths, candidate viewpoints, and target positions, facilitating qualitative analysis of system execution and 3D reconstruction progression.

\section{Experimental Setup and Framework Methodology} \label{Sec 3: res}

In this section, the sequential procedure followed by the pipeline is described, from the initial scanning stage of AgriSEE to obtaining the 3D position of the tomatoes.

\subsection{Virtual Environment Setup}

\begin{figure}[H]
    \centering
    \includegraphics[width=\linewidth]{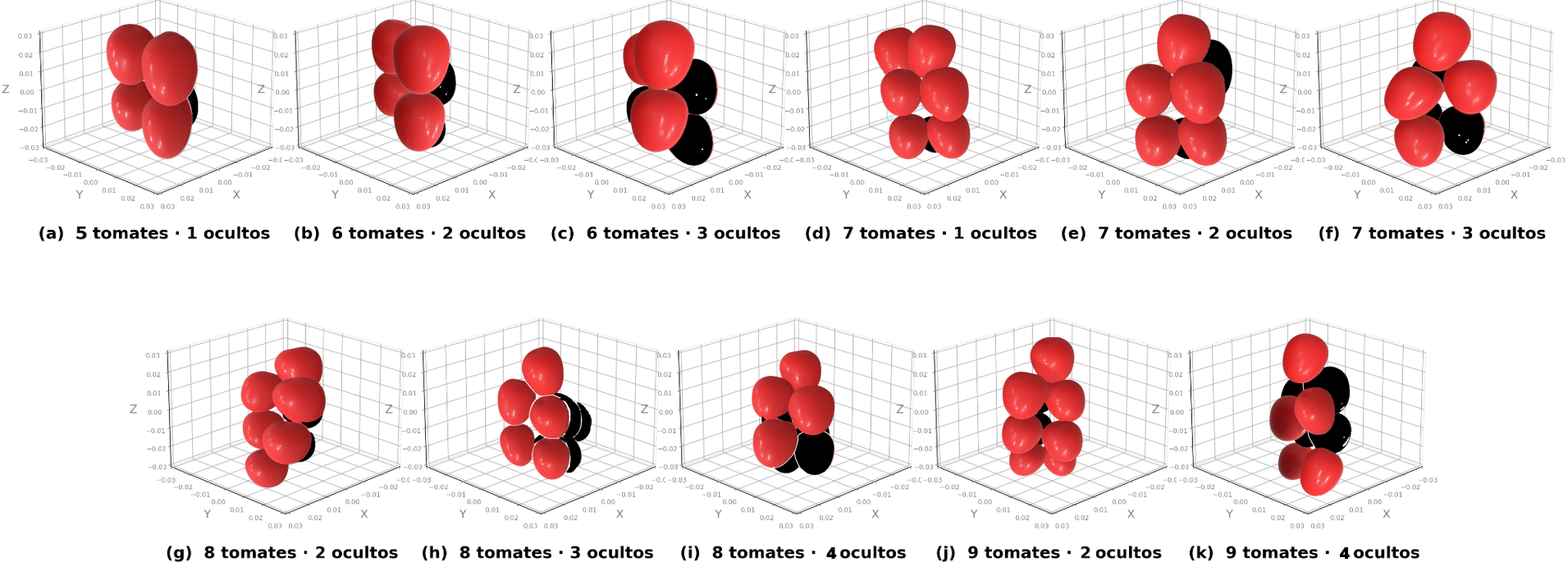}
    \caption{Geometry of the analyzed tomato clusters}
    \label{fig:11toma}
\end{figure}

On the one hand, the simulated environment was structured using the \texttt{URDF} files of the AGI mobile robot, the UR3 arm, and the L515 camera, defining a hierarchical kinematic chain. On the other hand, multiple tomato clusters with varying fruit counts were scanned, as illustrated in Fig.~\ref{fig:real_toma}, yielding a detailed three-dimensional reconstruction of each scene. A refined geometric model was then obtained by applying SOR filtering to remove stochastic noise, followed by RANSAC-based shape detection to accurately fit individual tomato geometries, resulting in a highly faithful 3D representation. The resulting models were validated against ground-truth, real-world measurements by manually assessing the scanned tomatoes in terms of size, relative position, occlusion state, and orientation. Fig.~\ref{fig:Figt2} shows one of the resulting 3D models of a tomato cluster, corresponding to plants reaching heights of up to 1.5~m; This validation was supported by data extracted from our previous work \citep{canadas2024greenbot}. To reduce computational overhead during simulation, this study considers a single tomato plant populated with 11 distinct cluster topologies, shown in Fig. \ref{fig:11toma}. To optimize computational cost, the scenario included a single tomato plant, although the tomato clusters composing it were substituted across the 11 experiments. 

\begin{figure}[H]
  \centering
  \begin{subfigure}{0.45\linewidth}\centering
    \includegraphics[width=4cm]{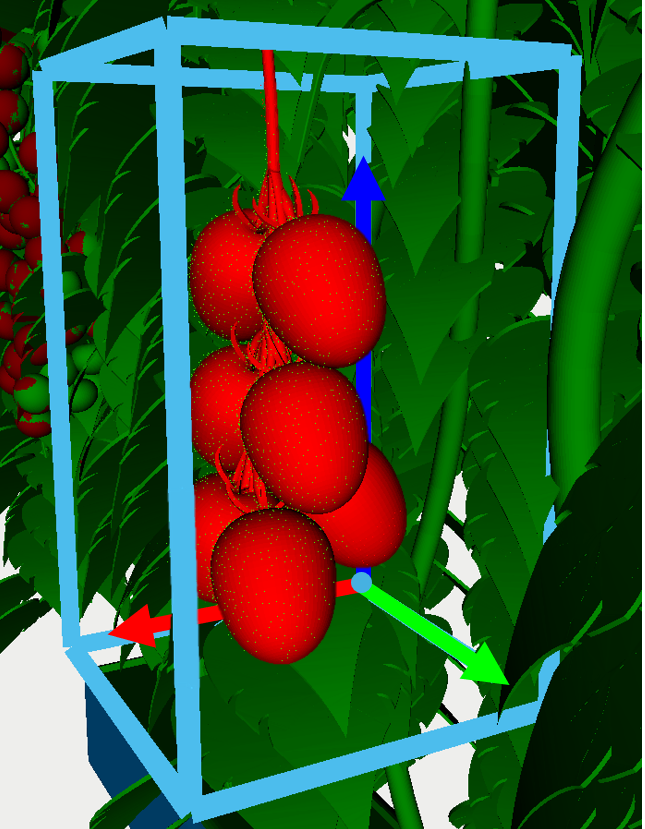}
    \caption{Scanning volume defined for data acquisition}
    \label{fig23}
  \end{subfigure}
  \begin{subfigure}{0.45\linewidth}\centering
    \includegraphics[width=4cm]{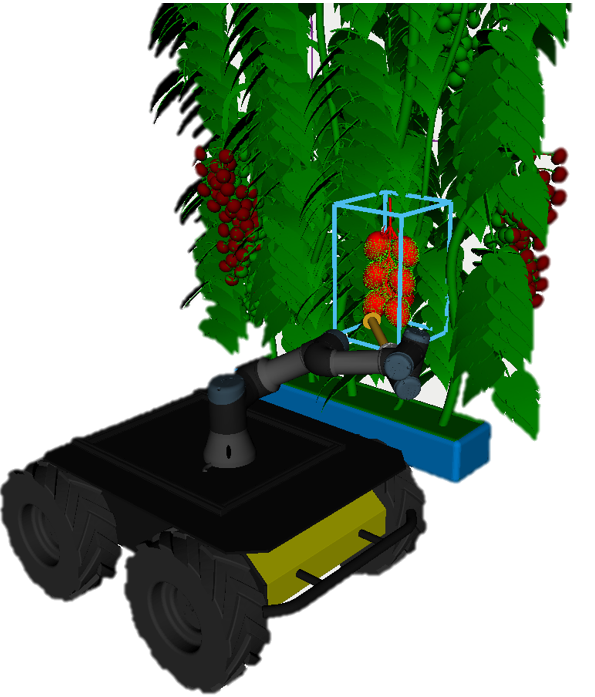}
    \caption{Scanned tomato cluster}
    \label{fig25}
  \end{subfigure}
  \caption{Setup configuration}
  \label{fig:Figt222}
\end{figure}

The scanning volume was defined such that the algorithm could correctly capture the partially occluded tomatoes present in the scene. This volume was specified by obtaining coherent measurements that encompass any tomato cluster in the greenhouse, positioning the robot at a distance expected in a real greenhouse scenario. During this definition, special attention was paid to minimizing interference with surrounding leaves, aimed at favoring the acquisition of relevant data on the fruit. In Fig. \ref{fig:real_toma}, it can be observed that, in this case, they do not usually contain an excessive amount of leaves and obstacles at the time of harvesting, which facilitates the use of this pipeline. Fig.~\ref{fig23} illustrates the scanning volume considered in the experiments. Once all \texttt{joint} elements were defined according to the established kinematic chain, the simulation environment shown in Fig.~\ref{fig25} was obtained.

To guarantee safe interaction during motion planning in ROS, several collision objects were defined in the \texttt{.SRDF} file. Specifically, a rectangular volume was implemented to delimit the space of the plant and protect the leaves, a concentric cylindrical volume around the cluster—designed to encompass various geometries and prevent the sensor from exceeding its minimum operating distance—and a platform on the top base of the AGI robot to prevent mechanical interference with the manipulator. This setup consolidated a functional environment suitable for executing the experiments.

\subsection{Pipeline Methodology}

Once the environment is defined, the 11 cluster models are evaluated (Fig.~\ref{fig:11toma}). The AgriSEE simulator incorporates a Gaussian noise model with $\sigma_{\text{noise}} = 0.01$\,m, nominal value from the manufacturer of the Intel RealSense L515 camera at the working distance employed (0.4--0.5\,m). For this reason, each cluster is scanned in 3 independent trials in order to empirically estimate the repeatability of the pipeline, calculating the standard deviation per axis of the estimated centroids across trials. Regarding the parameters used, Table \ref{tab:params_completos} shows the selected values.

\begin{table}[H]
\centering
\caption{Parameters used in the simulation}
\label{tab:params_completos}
\begin{tabular}{llll}
\hline
\textbf{Parameter} & \textbf{Symbol} & \textbf{Value} & \textbf{Units} \\
\hline
\multicolumn{4}{c}{\textbf{AgriSEE}} \\
\hline
Target density & $\rho$ & 2$\cdot$ $10^6$ & points/m$^{3}$ \\
Neighborhood radius & $r$ & 0.04 & m \\
Sensor - object distance & $d$ & 0.38 & m \\
Maximum number of views & $\tau$ & 50 & numbers of views \\
Implemented Gaussian noise & $\sigma_{\text{noise}}$ & 0.01 & - \\
\hline
\multicolumn{4}{c}{\textbf{SOR}} \\
\hline
Number of neighbors & $K$ & 20 & points \\
Threshold multiplier & $\alpha$ & 2.0 & - \\
\hline
\multicolumn{4}{c}{\textbf{RG}} \\
\hline
Search radius & $r_n$ & 0.006 & m \\
Maximum neighbors & $k_n$ & 25 & points \\
Neighborhood radius & $r_v$ & 0.006 & m \\
Angular threshold & $\theta_{\text{th}}$ & 12 & $^\circ$ \\
Minimum points & $N_{\min}$ & 350 & points \\
\hline
\multicolumn{4}{c}{\textbf{DBSCAN}} \\
\hline
Neighborhood radius (Epsilon) & $\varepsilon$ & 0.007 & m \\
Minimum core points & $N_{\min}^{\mathrm{DB}}$ & 60 & points \\
\hline
\end{tabular}%
\end{table}

\subsubsection{Cluster Scanning with AgriSEE}

Once the simulation environment was configured, the AgriSEE algorithm was configured with the \texttt{moveit} library, which loads the model into RViz to apply and solve the kinematic equations of the UR3. The parameters used are shown in Table \ref{tab:params_completos}, highlighting the observation distance $d$ of 0.38\,m and a radius $r$ of 0.04\,m due to the dimensions of the tomatoes. The rest of the parameters are kept fixed relative to those proposed. Algorithm \ref{alg:seepp} describes AgriSEE's view collection method.

\subsubsection{Noise and Outlier Filtering with SOR}

Once the scanning process was finalized, the resulting file in \texttt{.ply} format was processed using the SOR algorithm since, although this type of cluster does not feature a large amount of leaves, points derived from sensor noise or from the stem, leaves, or obstacles are usually obtained \citep{rusu20113d}. A number of neighbors $K$ of 20 is used to remove leaves along with a threshold of $2\sigma$. In this work, the filtering was not implemented in AgriSEE's \textit{C++} node, but rather separately, in order to evaluate each of the stages. Said adaptation will be carried out as future work. Algorithm \ref{alg:seepp} shows the combination of the filtering method with AgriSEE.

\begin{algorithm}[H]
\caption{AgriSEE + SOR Algorithm}
\label{alg:seepp}

\KwIn{Initial cloud $\mathcal{P}^0$ with radius $r$, distance $d$, threshold $\rho_{\min}$, SOR neighbors $K$, and multiplier $\alpha$}
\KwOut{Consolidated point cloud $\mathcal{P}$}

Initialize $k \leftarrow 0$\;
Acquire initial cloud $\mathcal{P}^0$\;

\While{$\max_{\mathbf{p}_i \in \mathcal{P}^k} \rho_i^k < \rho_{\min}$}{

    \For{each point $\mathbf{p}_i \in \mathcal{P}^k$}{
        Calculate local density $\rho_i^k$ within radius $r$\;
    }

    Classify points as core, boundary, or outliers according to $\rho_i^k$\;

    Identify boundary point set $\mathcal{F}^k$\;

    \For{each boundary point $\mathbf{p}_f \in \mathcal{F}^k$}{
        Estimate surface normal $\mathbf{e}_{\mathrm{n}}$\;

        \If{$\mathbf{e}_{\mathrm{n}} \cdot (\mathbf{p}_f - \mathbf{c}_o) > 0$}{
            Generate candidate view:\\
               $ \mathbf{v}_k = \mathbf{p}_f + d \, \mathbf{e}_{\mathrm{n}}$
        }
    }

    Select most informative view $\mathbf{v}_k$\;
    Move sensor to $\mathbf{v}_k$\;
    Acquire new partial cloud $\Delta \mathcal{P}^k$\;
    Integrate $\Delta \mathcal{P}^k$ into $\mathcal{P}^k$\;
    Filter $\mathcal{P}^k$ using SOR ($d_i \le \mu + \alpha \cdot \sigma$)\;

    $k \leftarrow k + 1$\;
}

Refine $\mathcal{P}$ using \textit{Iterative Closest Point (ICP)}\;
\Return $\mathcal{P}$\;
\end{algorithm}

\subsubsection{Cluster Segmentation with RG + DBSCAN}

Given that NBV reconstructions can generate incomplete and fragmented surfaces, especially in the presence of occlusions, RG is used to estimate local normals, calculated through a neighbor search within a fixed radius. Normals provide surface orientation information and constitute a fundamental geometric descriptor to distinguish between smooth regions belonging to the same object and discontinuities associated with changes in curvature or contacts between different objects. In the context of tomato clusters, this information is key to differentiating adjacent fruits even when they share physical contact zones. Following this process, DBSCAN is applied to obtain the clusters, which are assigned to tomatoes. Algorithm \ref{alg:tomato_segmentation} presents the proposed pseudocode. Likewise, to perform the classification, the parameters in Table \ref{tab:params_completos} were used.

\begin{algorithm}[H]
\caption{Segmentation via RG and DBSCAN}
\label{alg:tomato_segmentation}

\KwIn{$\mathcal{P}$, $r_v$, $\theta_{\text{th}}$, $N_{\min}$, and ($\varepsilon$, $N_{\min}^{\mathrm{DB}}$)}
\KwOut{$n_c$}

Remove initial outliers in cloud $\mathcal{P}$\;
Initialize: $n_c \leftarrow \emptyset$, $V \leftarrow \emptyset$, $k \leftarrow 0$\;

\For{each indexed point $p_i \in \mathcal{P}$}{
    \If{$p_i \notin V$}{
        Initialize $R \leftarrow \{p_i\}$ and $Q \leftarrow \{p_i\}$\;
        Register $p_i$ in the visited set $V$\;
        \While{$Q$ is not empty}{
            Extract: $p_j$ from $Q$, neighbors $\mathcal{N}(p_j)$ within $r_v$\;
            \For{each neighbor point $p_n \in \mathcal{N}(p_j)$}{
                \If{$p_n \notin V$}{
                    Calculate local angular deviation $\theta_{jn} = \arccos(\mathbf{n}_j \cdot \mathbf{n}_n)$\;
                    \If{$\theta_{jn} < \theta_{\text{th}}$}{
                        Add neighbor point $p_n$ to current region $R$\;
                        Register $p_n$ in the visited set $V$\;
                        Insert $p_n$ into processing queue $Q$\;
                    }
                }
            }
        }

        \If{$|R| \geq N_{\min}$}{
            Assign index label $k$ to all points forming region $R$ (Increment cluster counter $k \leftarrow k + 1$)\;
        }
    }
}

Isolate the unassigned residual point subset $\mathcal{U} \leftarrow \{ \mathbf{p}_i \in \mathcal{P} \mid \mathbf{p}_i \text{ does not belong to any cluster } R \}$\;

\If{$|\mathcal{U}| > N_{\text{hidden}}$}{
    Apply DBSCAN clustering exclusively to the residual subset $\mathcal{U}$\;
    \For{each resulting valid hidden dense cluster}{
        Assign a new unique instance label $k$ to the isolated segment\;
        Increment cluster counter $k \leftarrow k + 1$\;
    }
}
\Return Instance mapping matrix $n_c$\;
\end{algorithm}

\subsubsection{Localization}

Once all valid instances are identified, the geometric center of each cluster is calculated as the mean of its points, thus obtaining a compact representation of the estimated global localization of each tomato. This is possible because the AgriSEE algorithm saves, in a \texttt{csv} file, the view position relative to the origin of the UR3, the viewing direction of the L515 optical eye, and the observed point along the optical axis.

This allows knowing the exact location of the \textit{Tool Center Point} (TCP) during scanning. Finally, the positions of the tomatoes are transformed to the world frame $\{\mathcal{M}\}$, which acts as the base reference frame for all elements. From the points identified as tomatoes, the centroid of each instance is calculated by applying DBSCAN to the previously obtained region. This centroid, computed as the mean of the three-dimensional coordinates of the points composing each cluster, $\mathbf{c}_i \in \mathbb{R}^3$ with $i \in \{1,\dots,n_c\}$, provides a representative point that serves as an initial target for the robotic arm's approach during the future harvesting task, even if it does not coincide exactly with the geometric center of the tomato. The location of the cluster base reference frame $\{\mathcal{R}\}$ is defined at the lower-left corner of the background of the bounding box enclosing the cluster (Fig. \ref{fig23}). In this way, AgriSEE calculates the point cloud referred to the TCP at each view to subsequently perform a transformation of the estimated centroids $\mathbf{c}_i$ to the cluster base reference frame using the proposed pipeline. From each centroid, the principal growth direction of the fruit is additionally estimated, defined as the vector pointing from the centroid to the highest point in the point cloud $\mathcal{P}$. The estimated position of each tomato in homogeneous coordinates relative to $\{\mathcal{M}\}$ is obtained as:

\begin{equation}
\begin{bmatrix}
\mathbf{c}_{i}^{\mathcal{M}} \\
1
\end{bmatrix}
=
\mathbf{T}_{\mathcal{R}}^{\mathcal{M}}
\begin{bmatrix}
\mathbf{c}_i^{\mathcal{R}} \\
1
\end{bmatrix}.
\end{equation}

In particular, the location associated with each tomato is described by a reference frame origin at its estimated centroid, with its $z$-axis along the fruit growth direction, its $x$-axis pointing toward the region with the highest point density of the cluster, and its $y$-axis forming a right-handed system. This formulation allows defining a full pose for each detected fruit, providing the robotic system with the spatial information required to plan trajectories and execute harvesting tasks, as reflected in Algorithm \ref{alg:tomato_localization}.

\begin{algorithm}[H]
\caption{Estimated tomato localization in $\mathcal{M}$}
\label{alg:tomato_localization}

\KwIn{Point cloud $\mathcal{P}$, cluster $n_c\{i\}$}
\KwOut{Location $\{\mathbf{c}_i^{\text{cluster}}, \mathbf{c}_i^{\text{world}}\}$\;}

\For{each cluster $n_c$}{
    Calculate centroid w.r.t. cluster frame:
    \[
        \mathbf{c}_i^{\mathcal{R}} = \frac{1}{|C_i|} \sum_{\mathbf{p}_j \in C_i} \mathbf{p}_j
    \]
}

Read last NBV view from $\mathcal{C}$ and parameters\;

Homogeneous transformation $\mathbf{T}_{\mathcal{R}}^{\mathcal{M}}$\;

\For{each center $\mathbf{c}_i^{\text{cam}}$}{
    Transform to world frame:
    \[
        \mathbf{c}_i^{\mathcal{M}} =
        \left( \mathbf{T}_{\mathcal{R}}^{\mathcal{M}}
        \begin{bmatrix}
        \mathbf{c}_i^{\mathcal{R}} \\
        1
        \end{bmatrix}
        \right)_{1:3}
    \]
}

\Return $\{\mathbf{c}_i^{\text{cam}}, \mathbf{c}_i^{\mathcal{M}}\}$\;

\end{algorithm}

\section{Results}

In this section, the results of each stage are presented, and the metrics obtained from the eleven clusters used in the simulation are analyzed.

\subsection{AgriSEE Scanning and SOR Filtering}

\begin{figure}[H]
  \centering
  \begin{subfigure}{0.45\linewidth} \centering
    \includegraphics[width=5cm]{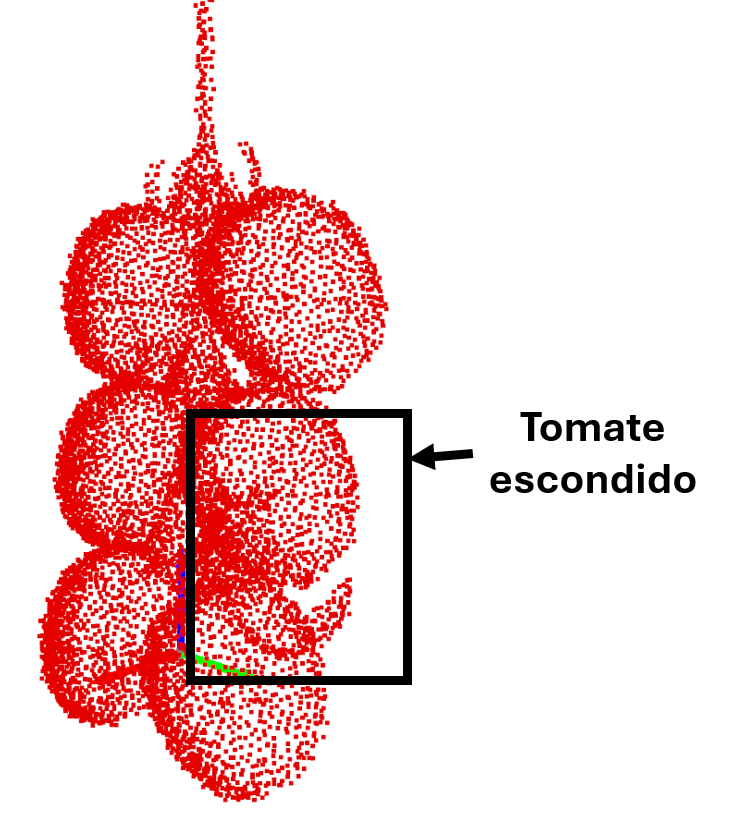}
    \caption{Isometric view}
    \label{fig:frontal2}
  \end{subfigure}
  \begin{subfigure}{0.45\linewidth} \centering
    \includegraphics[width=4.3cm]{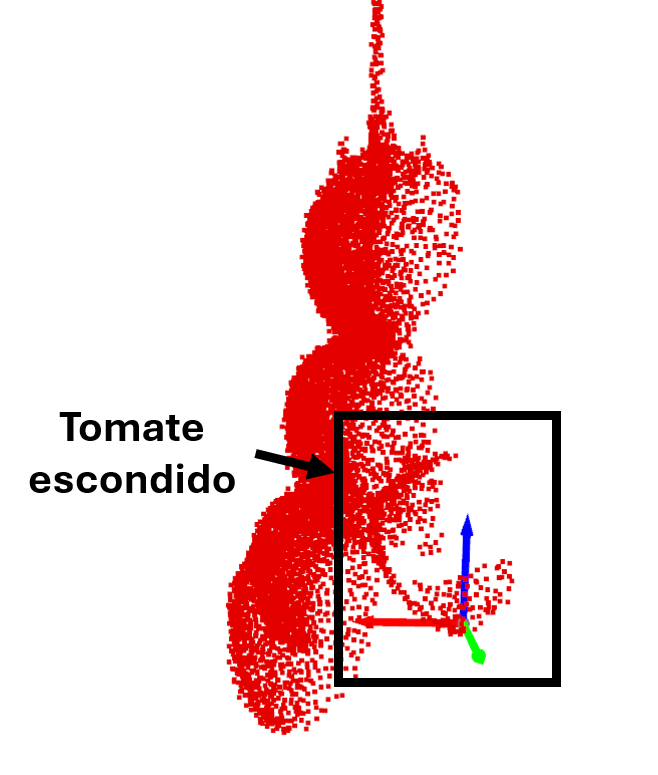}
    \caption{Side view}
    \label{fig:caballera}
  \end{subfigure}
  \caption{Three-dimensional model of the scanned tomato cluster.}
  \label{fig:Figt1}
\end{figure}

A total of 11 scans were performed on the tomato clusters shown in Fig. \ref{fig:11toma}, which were introduced into a volume similar to that in Fig. \ref{fig23}. Since the original \texttt{.STL} model is available, a comparison with the real ground truth can be performed to obtain metrics that serve to analyze the efficacy of the proposed pipeline. However, for illustrative purposes, the figures only display the 7-tomato model, of which only 1 is hidden and located at the bottom part. Fig. \ref{fig:Figt1} shows the result of the point cloud model scanned by AgriSEE and filtered with SOR. Likewise, Table \ref{tab:analisis_pipeline} presents the evaluation metrics of the simulations.

\begin{table}[H]
\centering
\caption{Analysis of AgriSEE scanning and SOR filtering}
\label{tab:analisis_pipeline}
\resizebox{\textwidth}{!}{%
\begin{tabular}{cccccccccc}
\hline
\textbf{Total} & \textbf{Hidden} & \textbf{No. of Views} & \textbf{No. of Points} & \textbf{No. of Points} & \textbf{Computation Time} & \textbf{Scanned} & \textbf{Precision} & \textbf{Recall} & \textbf{F1-Score} \\
\textbf{Tomatoes} & \textbf{Tomatoes} & \textbf{($\tau$)} & \textbf{post-AgriSEE} & \textbf{post-SOR} & \textbf{(s)}& \textbf{percentage (\%)} & \textbf{(\%)} & \textbf{(\%)} & \textbf{(\%)} \\
\hline
5 & 1 & 5  & 12460 & 7850  & 40.1 & 86.2 & 95.1 & 84.8 & 89.7 \\
6 & 2 & 7  & 12610 & 7620  & 46.8 & 82.8 & 93.8 & 81.2 & 87.0 \\
6 & 3 & 9  & 12330 & 7380  & 51.3 & 74.4 & 91.6 & 72.5 & 80.9 \\
\hline
7 & 1 & 12 & 14101 & 9240  & 72.1 & 80.6 & 95.3 & 79.1 & 86.4 \\
7 & 2 & 12 & 13940 & 9050  & 72.3 & 74.1 & 94.1 & 72.6 & 82.0 \\
7 & 3 & 13 & 13410 & 8810   & 76.1 & 65.9 & 92.4 & 64.2 & 75.7 \\
\hline
8 & 2 & 12 & 15742 & 10610 & 71.5 & 82.3 & 95.0 & 80.8 & 87.3 \\
8 & 3 & 13 & 15630 & 10240 & 77.2 & 71.5 & 92.9 & 69.8 & 79.7 \\
8 & 4 & 13 & 15112 & 10780 & 76.6 & 64.2 & 88.5 & 62.5 & 73.3 \\
\hline
9 & 2 & 14 & 17480 & 11950 & 82.3 & 71.4 & 95.3 & 69.9 & 80.6 \\
9 & 4 & 14 & 17220 & 11510 & 84.7 & 56.7 & 93.0 & 55.1 & 69.2 \\
\hline
\end{tabular}%
}
\end{table}

The SOR filtering stage effectively removes stochastic noise from the scene to consolidate clean clouds of between 7380 and 11,950 points without altering the fruit geometry, requiring computation times between 40.1\,s and 84.7\,s that scale linearly with the number of views ($\tau \in [5, 14]$). While precision remains high and stable across all scenarios (91.6\%--95.1\% in 6-tomato clusters), demonstrating the reliability of the algorithm against false positives, the recall degrades monotonically with increasing occlusions, reaching a minimum of 55.1\% in the most critical case of 9 tomatoes with 3 hidden ones. Increasing the number of viewpoints in dense configurations mitigates occlusions caused by frontal tomatoes; however, the loss of final completeness is due to a physical limitation imposed by the plant's blind spots rather than a deficiency in the filtering algorithm. Finally, a video\footnote{Simulation video: \url{https://youtu.be/FDF8YRBGnvM}} illustrating the complete data acquisition and processing workflow carried out by the algorithm has been made available.

\subsection{Segmentation with RG and DBSCAN}

The point cloud resulting from filtering is used as input to RG and DBSCAN. The obtained regions and their classification into clusters are shown in Fig. \ref{fig:loc}, while the evaluation metrics are presented in Table \ref{tab:analisis_segmentacion}.

After the RG stage, out of the 8,325 points in the filtered cloud, approximately 2,632 are assigned to the 7 detected clusters ($\sim$439 points per instance), whereas the remaining 5,693 points ($\sim$68.4\%) constitute the unsegmented residual transferred to the DBSCAN stage for recovering occluded instances. The method achieves $P_{\text{obj}} = 100\%$ under all conditions, as all regions are located on tomatoes. Recovery degrades gradually with occlusion, dropping from 100\% with one hidden fruit to 50\% under extreme occlusion conditions (5 out of 8 hidden fruits), resulting in an average $R_{\text{obj}}$ of 82.8\%. The segmentation quality ($mIoU$) follows the same trend, ranging between 88.4\% and 67.2\%, with an average of 80.7\%.

\begin{table}[H]
\centering
\caption{RG and DBSCAN Evaluation}
\label{tab:analisis_segmentacion}
\resizebox{\textwidth}{!}{%
\begin{tabular}{cccccccc}
\hline
\textbf{Total} & \textbf{Hidden} & \textbf{Real Tomatoes} & \textbf{Detected Tomatoes} & \textbf{Counting Error} & \textbf{Obj. Precision} & \textbf{Obj. Recall} & \textbf{Seg. Quality} \\
\textbf{Tomatoes} & \textbf{Tomatoes} & \textbf{($N_{\text{real}}$)} & \textbf{($N_{\text{det}}$)} & \textbf{($CE$)} & \textbf{($P_{\text{obj}}$ \%)} & \textbf{($R_{\text{obj}}$ \%)} & \textbf{($mIoU$ \%)} \\
\hline
5 & 1 & 6 & 6 & 0 & 100.0 & 100.0 & 88.4 \\
6 & 2 & 6 & 5 & 1 & 100.0 & 83.3  & 85.1 \\
6 & 3 & 6 & 4 & 2 & 100.0 & 66.7  & 79.2 \\
\hline
7 & 1 & 7 & 7 & 0 & 100.0 & 100.0 & 87.3 \\
7 & 2 & 7 & 6 & 1 & 100.0 & 85.7  & 83.9 \\
7 & 3 & 7 & 5 & 2 & 100.0 & 71.4  & 76.5 \\
\hline
8 & 2 & 8 & 8 & 0 & 100.0 & 100.0 & 86.8 \\
8 & 3 & 8 & 6 & 2 & 100.0 & 75.0  & 78.3 \\
8 & 4 & 8 & 4 & 4 & 100.0 & 50.0  & 67.2 \\
\hline
9 & 2 & 9 & 9 & 0 & 100.0 & 100.0 & 85.9 \\
9 & 4 & 9 & 7 & 2 & 100.0 & 77.8  & 77.6 \\
\hline
\multicolumn{3}{l}{\textbf{Average}} & \textbf{-} & \textbf{1.3} & \textbf{100.0} & \textbf{82.7} & \textbf{81.5} \\
\hline
\end{tabular}%
}
\end{table}

\subsection{Localization Estimation}

\begin{figure}[H]
    \centering
    \includegraphics[width=0.55\linewidth]{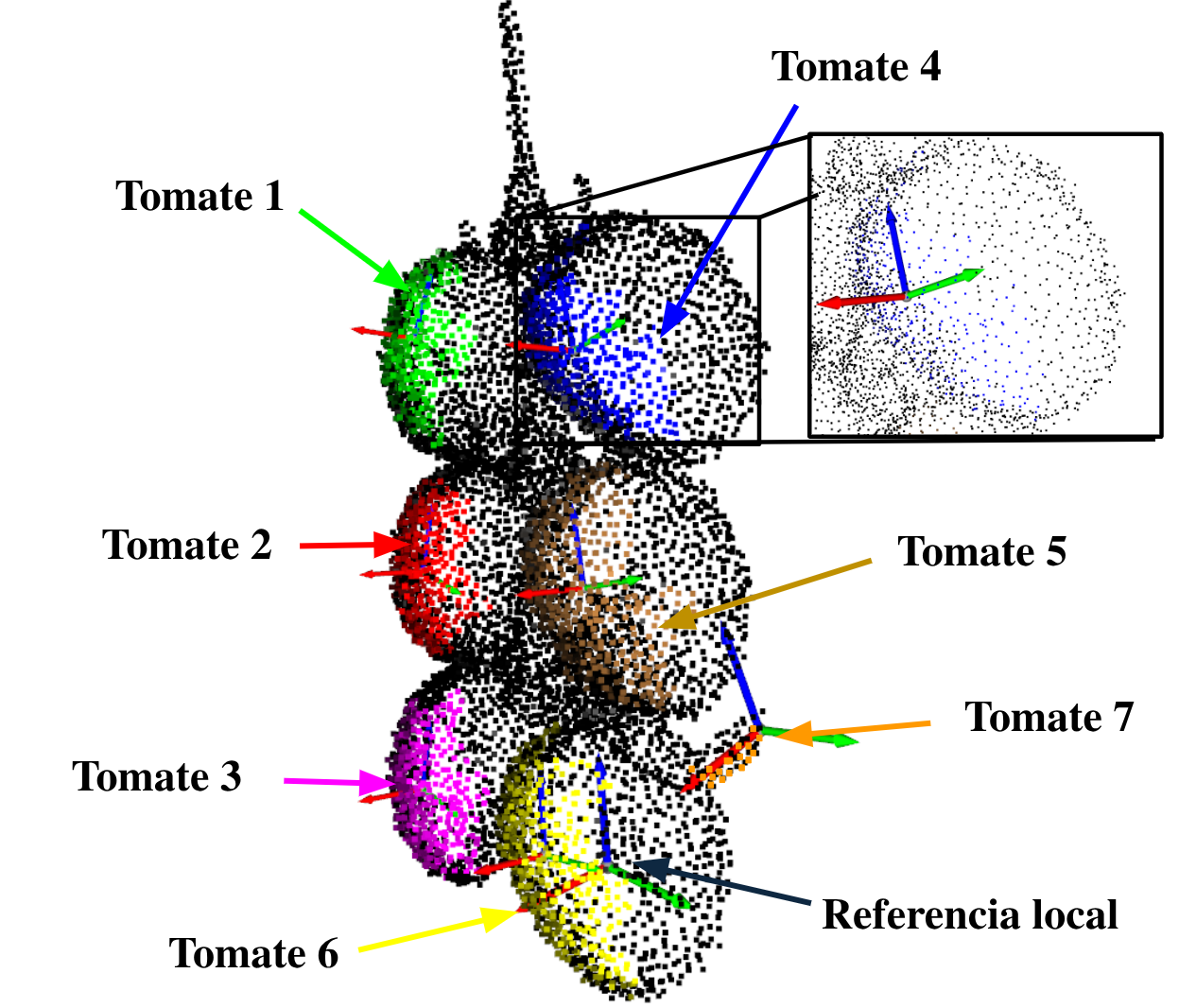}
        \caption{Tomato segmentation using RG + DBSCAN and local localization of each segmented tomato. Yellow, brown, red, blue, magenta, and green represent the frontal tomatoes, while orange represents the hidden tomato.}
    \label{fig:loc}
\end{figure}

Fig.~\ref{fig:loc} illustrates the resulting localization of each tomato in the workspace, showing the estimated centroids and the orientation associated with the growth direction of the fruit with respect to the structure of the tomato plant. Table~\ref{tab:tomato_centers_uncertainty} presents the estimated geometric centers of the detected tomatoes, expressed both in the cluster and world reference frames, computed from the NBV reconstruction.

The reported centroids correspond to the centroid of the segmented point set, not the actual geometric center of the fruit; this approximation is sufficient to guide the robotic arm during the approach phase, delegating the final gripper refinement to RGB visual feedback. The sensor positions (0.4--0.5\,m from the cluster) and observation vectors (negative X-axis, unit magnitude) are consistent with the planned scanning configuration. In the TCP frame, the centroids are concentrated within a volume of a few centimeters and exhibit a progressive increase in $Z$ for higher fruits, correctly reflecting the vertical structure of the cluster. Overall, the pipeline provides spatially consistent estimations suitable for approach and harvesting motion planning tasks.

In agricultural production systems of the Mediterranean environment, plants undergo continuous vertical guiding and trellising, a commercial practice that displaces the clusters outward from the plant structure to facilitate manual harvesting. This spatial distribution drastically reduces the presence of natural obstacles in the immediate environment of the fruit, thereby decreasing the probability of the algorithm incurring false positives due to geometric confusion with other plant organs.

\begin{table}[H]
\centering
\caption{Geometric centers of the detected tomatoes in cluster and world coordinates. Uncertainty ($\sigma$) reflects the standard deviation of points within each detected cluster. The RMSE column indicates centroid stability across 3 independent trials (system repeatability).}
\label{tab:tomato_centers_uncertainty}
\resizebox{\textwidth}{!}{%
\begin{tabular}{c ccc ccc cc}
\hline
\multirow{2}{*}{\textbf{Tom.}} &
\multicolumn{3}{c}{\textbf{Cluster Center (m)}} &
\multicolumn{3}{c}{\textbf{World Center (m)}} &
\multirow{2}{*}{\textbf{RMSE}} \\
\cline{2-7}
 & $x \pm \sigma_x$ & $y \pm \sigma_y$ & $z \pm \sigma_z$ &
   $x \pm \sigma_x$ & $y \pm \sigma_y$ & $z \pm \sigma_z$ & (mm) \\
\hline
1 & $0.065 \pm 0.006$ & $0.044 \pm 0.007$ & $0.044 \pm 0.008$ &
    $0.060 \pm 0.007$ & $0.519 \pm 0.009$ & $0.154 \pm 0.010$ & 3.6 \\

2 & $-0.032 \pm 0.009$ & $0.039 \pm 0.011$ & $0.043 \pm 0.012$ &
    $-0.023 \pm 0.011$ & $0.521 \pm 0.013$ & $0.154 \pm 0.014$ & 4.1 \\

3 & $0.051 \pm 0.012$ & $0.047 \pm 0.015$ & $0.133 \pm 0.016$ &
    $0.064 \pm 0.014$ & $0.555 \pm 0.017$ & $0.238 \pm 0.018$ & 4.7 \\

4 & $-0.046 \pm 0.007$ & $0.050 \pm 0.009$ & $0.129 \pm 0.010$ &
    $-0.033 \pm 0.008$ & $0.550 \pm 0.011$ & $0.236 \pm 0.012$ & 3.9 \\

5 & $0.052 \pm 0.015$ & $0.042 \pm 0.019$ & $0.214 \pm 0.021$ &
    $0.060 \pm 0.017$ & $0.577 \pm 0.022$ & $0.318 \pm 0.023$ & 5.2 \\

6 & $-0.037 \pm 0.010$ & $0.053 \pm 0.012$ & $0.226 \pm 0.013$ &
    $-0.035 \pm 0.011$ & $0.573 \pm 0.014$ & $0.331 \pm 0.015$ & 4.3 \\

7 & $0.032 \pm 0.018$ & $0.002 \pm 0.022$ & $0.085 \pm 0.024$ &
    $0.030 \pm 0.020$ & $0.536 \pm 0.025$ & $0.087 \pm 0.026$ & 5.6 \\

\hline
\textbf{Average} & & & & & & & \textbf{4.5} \\
\hline
\end{tabular}%
}
\end{table}

The average RMSE of 4.5~mm obtained across the seven tomatoes is well within the tolerance required for robotic harvesting. Considering that the gripping mechanism of the robotic end-effector, together with the compliant nature of the fruit surface, typically accommodates positioning deviations on the order of several millimeters to a few centimeters, this level of centroid accuracy is more than sufficient to guarantee a successful and reliable picking operation. Consequently, the observed error does not compromise the feasibility of autonomous harvesting, but rather confirms the suitability of the proposed pipeline for practical deployment \citep{carbone2025development}.

\section{Conclusion} \label{Sec 4: conc}

In this work, a comprehensive approach for the 3D reconstruction, segmentation, and localization of fruits in greenhouse environments has been presented, aimed at automated harvesting applications. Starting from the inherent limitations of traditional perception systems in agricultural scenarios—such as tomato occlusions, geometric variability of plants, and adverse environmental conditions—it has been demonstrated that combining active perception through NBV planning with DBSCAN segmentation strategies constitutes an effective and robust solution for the individual characterization of fruits in dense clusters.

The obtained results confirm that the use of the AgriSEE algorithm, integrated within the SEE++ system, enables the generation of high-quality 3D reconstructions from incremental observations. The SOR filtering stage successfully removes simulated sensor noise, consolidating point clouds with a geometric accuracy consistently remaining above 91.6\%, without requiring prior models or assumptions regarding the complete geometry of the object. Furthermore, it has been demonstrated that the reduction in overall recall under severe occlusion configurations corresponds to the physical visibility constraints imposed by the environment (reaching 55.1\% in the worst evaluated scenario), validating a solid algorithmic performance whose processing times scale predictably with the number of views ($\tau$).

Moreover, the proposed segmentation process, based on RG guided by geometric continuity and complemented with DBSCAN clustering, has resolved one of the most complex problems in clustered fruits. As a result, it was possible to isolate individual fruits, achieving a perfect object precision ($P_{\text{obj}}$) of 100\% and an average segmentation quality ($mIoU$) exceeding 80\%. Additionally, the correct spatial localization of each fruit has been validated through the estimation of its centroids in the cluster reference frame and its subsequent transformation to the world frame. The observed geometric consistency among sensor positions, view orientation, and spatial distribution of the tomatoes confirms its validity for guiding robotic arm motion planning. Being a sequential architecture specifically designed around individual methods, establishing a performance comparison with isolated global algorithms (such as standard DBSCAN or HDBSCAN) was not deemed necessary, as the value of the proposal lies in the synergy of the full pipeline rather than the optimization of a single classifier.

However, the proposed solution presents certain practical limitations that must be considered to evaluate its viability in real-world deployments. The evaluated NBV planning requires capturing up to 14 views for a single cluster, consuming a computation time of approximately 84 seconds in the worst-case scenario. This temporal metric suggests that sequentially scaling across all clusters of an entire plant would be extremely costly, limiting the temporal throughput required in the agricultural industry. Furthermore, the transition to the physical environment will require a paradigm shift in evaluation; lacking an exact 3D ground truth (CAD model) in production greenhouses, the system will not be able to measure recall or IoU in real time, requiring the design of new indirect completeness metrics based on estimated volumetric densities. It should be noted, however, that the 3D tomato cluster models underlying the simulated scenarios were not synthetically generated, but derived from real greenhouse data and previously validated against manual ground-truth measurements of fruit size, centroid position, and orientation, which reinforces the realism of the simulated evaluation despite the current lack of a real-time ground truth in production settings. Consequently, before deploying the robot fully autonomously, establishing a supervised validation stage to monitor real-sensor responses under variable lighting will be essential.

As future lines of research, implementing all stages within a single \textit{C++} node so that AgriSEE executes the pipeline is proposed. Furthermore, evaluating the scalability and automation of the system against a high volume of clusters distributed simultaneously throughout the plant—optimizing global cycle times as well as extending to other plant types—remains a priority. To address the temporal limitations of planning, the implementation of an adaptive search space will be explored to dynamically reduce the required number of views ($\tau$) without compromising the geometric completeness of the exploration. Likewise, experimental validation of the system in real greenhouse environments under variable lighting conditions is planned; in this regard, the impact of direct solar radiation on the LiDAR sensor will be evaluated and, if necessary, the integration of physical shielding systems on the end-effector or the use of sensors with higher immunity to the ambient infrared spectrum will be considered. Additionally, future work will address direct integration with real-time grasp planners and controllers to execute physical fruit harvesting. Finally, extending this approach to other horticultural crops with similar morphologies will enable progress toward robotic platforms capable not only of automating harvesting but also of providing accurate yield estimations across the entire crop cycle.

\section*{Acknowledgments} 
This work has been carried out within the framework of the LIFE-ACCLI\\AMTE project (LIFE23-CCAES-LIFE-ACCLIMATE/101157315), and a Er-\\asmus+ Mobility grants from CeiA3. The first author, Fernando Cañadas-Aránega, holds an FPI grant (PRE2022-102415) from the Spanish Ministry of Science, Innovation, and Universities. The authors would like to express their gratitude to Prof. Dr. Margarita Chli, who supervised Fernando's research stay at the University of Cyprus (UCY), Cyprus.
\bibliographystyle{cas-model2-names}

\bibliography{cas-refs}




\end{document}